\documentclass[letterpaper]{article} 
\usepackage{aaai2027}  
\usepackage[hyphens]{url}  
\usepackage{graphicx} 
\usepackage{natbib}  
\usepackage{caption} 
\usepackage{algorithm}
\usepackage{algorithmic}

\usepackage{tcolorbox}
\usepackage{xcolor}
\tcbuselibrary{breakable}
\usepackage{longtable, tabularx}
\usepackage{array}   
\usepackage{enumitem} 
\usepackage{caption}

\usepackage{listings}

\usepackage{float}
\usepackage{amsmath}
\usepackage{mathtools}
\usepackage{amsthm}
\usepackage{multirow}
\usepackage{colortbl}
\usepackage{xcolor}
\usepackage{tcolorbox}
\usepackage{xcolor}
\usepackage{amsfonts}
\usepackage{xspace}
\definecolor{base1024}{HTML}{57bb8a} 
\definecolor{base256}{HTML}{b5c975}  
\definecolor{base64}{HTML}{ffc964}   
\definecolor{base16}{HTML}{ee7c70}   
\colorlet{budget1024}{base1024!10}
\colorlet{budget256}{base256!10}
\colorlet{budget64}{base64!10}
\colorlet{budget16}{base16!10}
\definecolor{headerGray}{gray}{0.9}    
\definecolor{posGreen}{HTML}{009900} 
\usepackage{amsfonts}

\newcommand{\memo}{MEMO\xspace}

\usepackage{pifont}

\usepackage{newfloat}
\usepackage{listings}
\DeclareCaptionStyle{ruled}{labelfont=normalfont,labelsep=colon,strut=off} 
\floatstyle{ruled}
\newfloat{listing}{tb}{lst}{}
\floatname{listing}{Listing}

\usepackage{booktabs}
\nocopyright
\usepackage{newfloat}
\usepackage{listings}
\DeclareCaptionStyle{ruled}{labelfont=normalfont,labelsep=colon,strut=off} 
\floatstyle{ruled}
\newfloat{listing}{tb}{lst}{}
\floatname{listing}{Listing}

\usepackage{booktabs}

\title{MEMO: Multimodal Evidence Memory Organization for Long-Horizon LLM Agents}

\author{
    Xian Gao\textsuperscript{\rm 1}\equalcontrib,
    Jinpeng Wang\textsuperscript{\rm 2}\equalcontrib,
    Jiacheng Ruan\textsuperscript{\rm 1},
    Guangyu Cao\textsuperscript{\rm 3},
    Ting Liu\textsuperscript{\rm 1}\corresponding, and 
    Yuzhuo Fu\textsuperscript{\rm 1}\corresponding
}
\affiliations{
    \textsuperscript{\rm 1}Shanghai Jiao Tong University, 
    \textsuperscript{\rm 2}Tsinghua University, \\
    \textsuperscript{\rm 3}Institute of Automation of the Chinese Academy of Sciences

}

\begin{document}

\maketitle
\begin{abstract}
Long-running LLM agents rely on external memory to store and reuse information beyond a single context window, yet there is a fundamental tension between the continuous accumulation of interaction trajectories and the limited context capacity. The key challenge in agent memory is therefore not only to retrieve relevant records, but also to select necessary evidence under a given budget and organize it in an appropriate modality. Existing memory readout methods mainly use textual or visual forms. Text preserves high fidelity, but its linear token representation makes contents with different importance compete for the limited context at nearly uniform unit cost. Visual readout renders text into document-like images, which can use two-dimensional layouts to expose structure and emphasize key information, but it may lose fine-grained details during rendering and compression. To address this issue, we propose \textbf{MEMO}, a \text{M}ultimodal \textbf{E}vidence \textbf{M}emory \textbf{O}rganization method for LLM agents. MEMO first uses a trained evidence extractor to select relevant memory blocks and form evidence units with source information and presentation requirements. A trained query-conditioned memory manager assigns each unit to a textual, visual, or dual-channel carrier and selects a layout that matches the evidence structure. A deterministic memory construction module then generates the textual package and visual pages. The memory manager is trained with feedback from an offline reader that measures the utility of the guided memory plan, so that retention and presentation decisions align with downstream usage. We evaluate MEMO on four benchmarks, HotpotQA, 2WikiMultiHopQA, LoCoMo, and ALFWorld, with multiple reader backends. The results show that MEMO presents memory more efficiently with fewer memory tokens, improves downstream task performance, and builds more effective working memory under constrained budgets. For example, under a fixed 128-token budget with Qwen3-VL-32B as the reader, MEMO achieves an F1 score of 73.91 on 2Wiki, outperforming text-only memory with 56.26 and visual-only memory with 35.89.

\end{abstract}

\begin{figure}
    \centering
    \includegraphics[width=\linewidth]{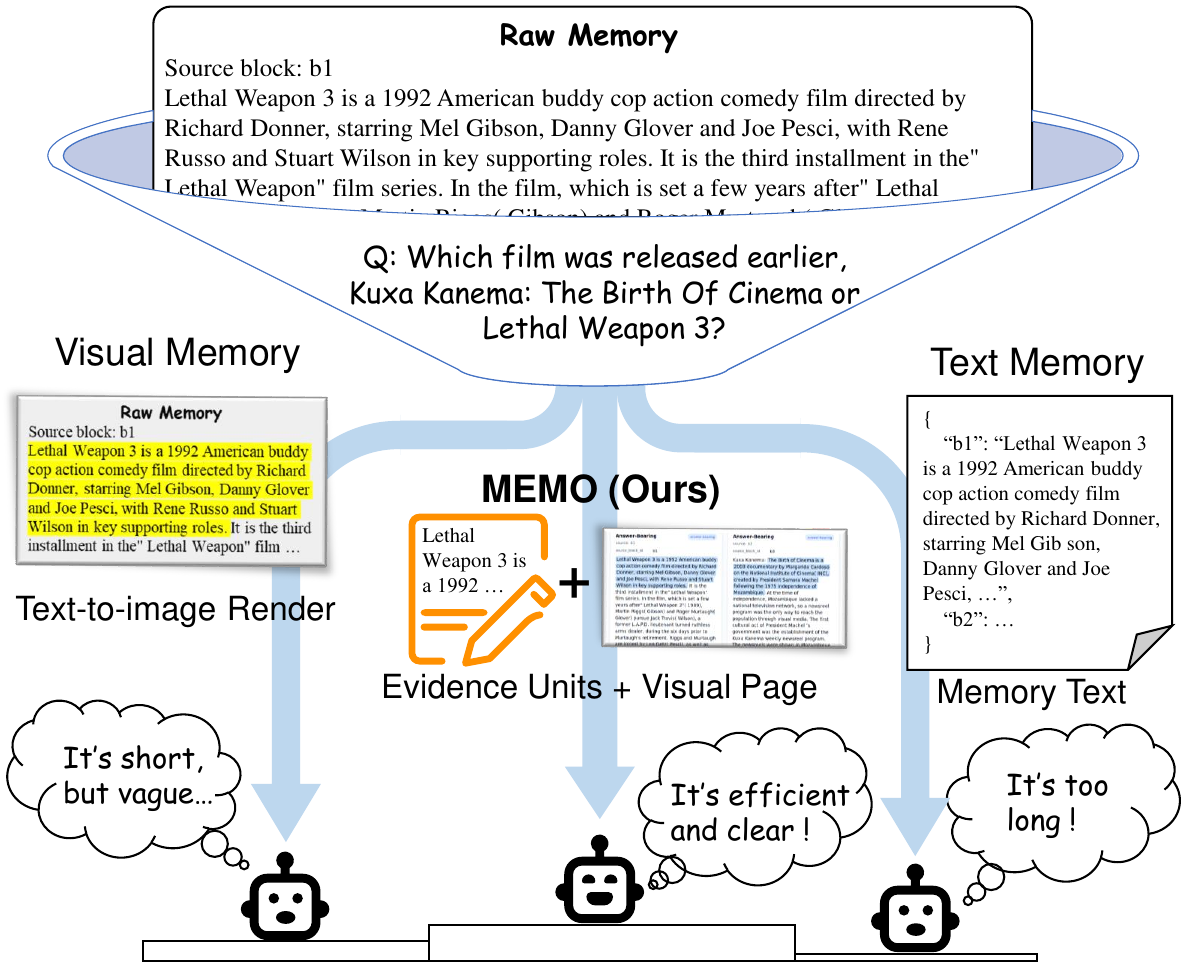}
    \caption{Motivation of MEMO.}
    \label{fig:head}
\end{figure}

\section{Introduction}

Large language models (LLMs) are increasingly evolving into agents that continuously perceive their environments, invoke tools, and execute complex tasks. In long-horizon interactions, agents need external memory to store information beyond a single context window and to recall relevant records in later actions. However, as interaction trajectories accumulate, memory grows continuously. Feeding the full history directly into the model is not only costly, but also forces truly useful evidence to compete with a large amount of redundant or weakly relevant content for the limited context. The central challenge of long-term memory therefore goes beyond storage and retrieval. It also lies in memory readout: given the current query, candidate records, and a context budget, the memory system needs to select the evidence necessary for the task and organize it into a working memory with clear structure that the model can read efficiently.

As illustrated in Figure \ref{fig:head}, existing methods implement memory readout mainly in textual or visual form. Textual memory systems build compact contexts through segmentation, compression, and retrieval. For example, SimpleMem and LightMem compress and retrieve historical records; MemRL, Agentic Memory, and Mem-$\alpha$ learn record utility or memory operations; EverMemOS and MemoryOS further manage memory hierarchies and life cycles. However, their final readout is still represented as a linear sequence of text tokens, which makes contents with different importance compete for context space at nearly uniform unit cost. Visual memory offers a complementary solution. MemOCR renders textual memory into visual pages with layout structure, while AgentOCR encodes agent histories into OCR-style images through optical self-compression. Image rendering can highlight important evidence by emphasizing text spans and can represent information through page layouts, thereby reducing the number of tokens needed to express the same content. However, after text is rendered into images, fine-grained semantic information may be difficult to preserve accurately. These observations suggest that textual and visual representations are suited to different types of evidence. Restricting all memory to a single modality cannot simultaneously achieve information fidelity, structural expressiveness, and resource efficiency.

To address this issue, we propose MEMO, a multimodal evidence-based memory organization method. Its core idea is to jointly decide what to preserve and how to present it at the granularity of evidence units, rather than converting retrieved results into a single modality. Specifically, the trained evidence extractor in \memo reads the query and candidate records, selects relevant source blocks, and predicts character-level spans that contain key facts. The selected spans and the original records together form traceable evidence units. Each unit preserves its source, specifies which details must be kept exactly, and indicates whether visual structures such as tables, timelines, lists, or cards can improve readability. This design protects key semantic information and provides compact operational objects for subsequent management. For each evidence unit, the trained memory manager jointly selects a text, image, dual-channel, or deletion action under a unified resource budget. The text representation preserves precise semantics, the visual representation presents structural relations through different layouts, and the dual-channel representation supplements visual structure while retaining textual anchors for semantic precision. The memory manager is trained with offline evaluations of candidate memory plans by a frozen reader model, which aligns evidence selection and presentation decisions with downstream task utility. Finally, the working memory building process constructs the working memory from the evidence units and their presentation actions, and verifies the total resource usage of textual and visual content.

We evaluate MEMO on HotpotQA \cite{yangHotpotQADatasetDiverse2018}, 2WikiMultiHopQA \cite{hoConstructingMultihopQA2020}, LoCoMo \cite{maharanaEvaluatingVeryLongTerm2024}, and ALFWorld \cite{shridharALFWorldAligningText2020}, covering multi-hop document question answering, long-term dialogue memory, and embodied interaction trajectory memory. We use multiple frozen VLMs as backend reader models. The experimental results show that MEMO achieves the best performance across all three reader models considered in the paper, both under a fixed 128-token budget and without a token budget. This indicates that evidence-level modality assignment and layout organization improve the use of limited context. The advantage is especially clear in multi-hop document question answering. With Qwen3-VL-32B as the reader model, MEMO achieves an F1 score of 73.91 on 2Wiki, while purely textual and purely visual memory achieve F1 scores of 56.26 and 35.89, respectively.

This paper makes three main contributions. First, we formulate agent memory readout as query-conditioned working memory construction under a shared resource budget, integrating evidence selection, presentation modality assignment, and visual organization into a unified optimization framework. Second, we propose MEMO, a multimodal memory organization method centered on traceable evidence units. By combining learned evidence extraction, memory planning, and deterministic working memory construction, MEMO exploits the complementarity between textual and visual structures, reducing the number of tokens required for memory while preserving semantic information. Third, we systematically evaluate MEMO across diverse memory sources and reader models, covering downstream task performance, resource usage, and adaptation across readers. The results validate the effectiveness of MEMO for long-horizon agent memory organization.
\section{Related Work}

\subsection{Textual External Memory}

Textual external memory is the most common memory interface for long-horizon LLM agents. Recent systems learn when to store, retrieve, revise, and compress records. MemRL estimates memory utility through runtime reinforcement learning over episodic memory \cite{zhangMemRLSelfEvolvingAgents2026}. Agentic Memory places short-term and long-term memory operations in the agent action space \cite{yuAgenticMemoryLearning2026}. SimpleMem and LightMem emphasize compact memory construction, topic-aware segmentation, and efficient retrieval \cite{liuSimpleMemEfficientLifelong2026,fangLightMemLightweightEfficient2025}. Mem-$\alpha$ acquires memory construction policies through reinforcement learning and provides a competitive trained text-memory baseline \cite{wangMemaLearningMemory2025}. These methods primarily study how memory content is formed and recalled. \memo extends this line to the read interface with a trained memory manager that selects evidence units and organizes their presentation for the final reader.

\subsection{Structured and Lifecycle-Aware Memory}

Another line of work models memory as a managed system with explicit structure and lifecycle operations. EverMemOS organizes episodic traces into higher-level memory objects with lifecycle control \cite{huEverMemOSSelfOrganizingMemory2026}. MemoryOS separates short-term, medium-term, and long-term memory through migration rules \cite{kangMemoryOSAI2025}. Graph and event memory systems such as MAGMA, SYNAPSE, ES-Mem, CAM, A-MEM, and Amory show that temporal, causal, entity, event, and narrative relations can support long-horizon reasoning \cite{jiangMAGMAMultiGraphBased2026,jiangSYNAPSEEmpoweringLLM2026,zouESMemEventSegmentationBased2026,liCAMConstructivistView2025,xuAMEMAgenticMemory2025,zhouAmoryBuildingCoherent2026}. \memo can consume structured records produced by these systems because evidence units preserve provenance, roles, priorities, and presentation requirements. It constructs working memory during the read stage so that structure in long-term storage can be used in a form suited to the current reader.

\subsection{Visual and Optical Memory}

Visual memory methods organize memory as two-dimensional artifacts. MemOCR converts textual memory into layout-aware visual pages and uses budget-aware learning to support long-horizon reasoning \cite{shiMemOCRLayoutAwareVisual2026}. AgentOCR represents agent history as OCR-style images through optical self-compression \cite{fengAgentOCRReimaginingAgent2026}. These studies show that images can carry memory content and that readers are sensitive to visual organization. \memo places presentation-form selection inside unit-level management so the system can retain a textual precision anchor, use visual structure, and choose dual presentation when a unit needs both kinds of information.

\subsection{Multi-Agent Memory and Context Routing}

Multi-controller memory systems assign management operations to specialized modules. MIRIX coordinates multiple memory components for LLM agents \cite{wangMIRIXMultiAgentMemory2025}. G-Memory builds hierarchical graph memory for multi-agent systems \cite{zhangGMemoryTracingHierarchical2025}. AMA uses collaborative agents for adaptive memory \cite{huangAMAAdaptiveMemory2026}. RCR-Router studies role-aware context selection under limited resources \cite{liuRCRRouterEfficientRoleAware2025}. These studies demonstrate the value of explicitly managing model input. \memo trains a read policy at evidence-unit granularity and uses frozen-reader outcomes to evaluate the final text and visual memory artifact.

\begin{figure*}
    \centering
    \includegraphics[width=0.98\linewidth]{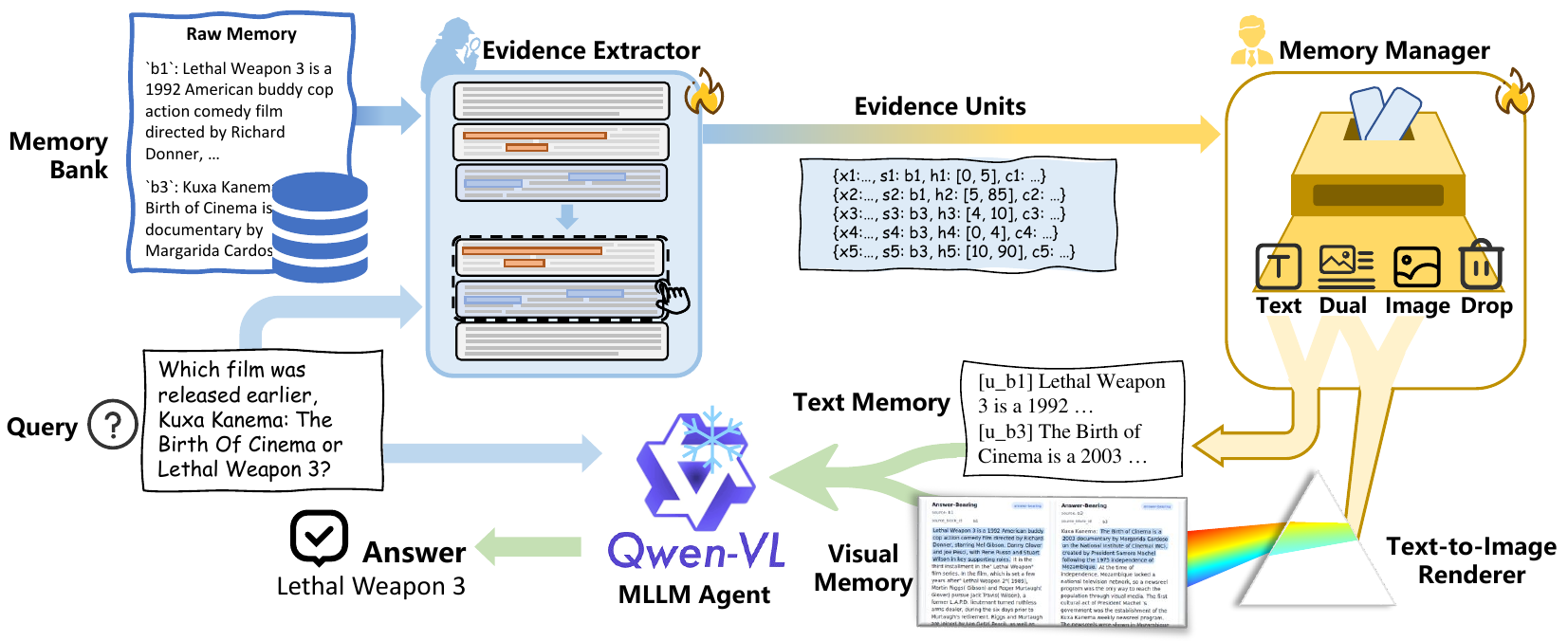}
    \caption{\textbf{Workflow of MEMO}. MEMO extracts evidence units from raw memory, plans their presentation as text, image, dual-channel, or dropped content, and builds a query-specific working memory for the multimodal LLM agent.}
    \label{fig:workflow}
\end{figure*}

\section{Method}

\subsection{Task Definition and Overview}
Figure \ref{fig:workflow} illustrates the workflow of MEMO. Given a current query $q$, memory records $\mathcal{R}$, and a total memory budget $B$, \memo constructs working memory $a=(T,V)$ for one reader call, where $T$ is a text packet and $V$ is an optional visual page. A trained evidence extractor first identifies source blocks relevant to the query and predicts character-offset highlights for key facts. A deterministic unit-materialization step copies the corresponding content from the original records, preserves its source links and highlights, and produces a set of evidence units $\mathcal{U}$. The memory manager then decides which units enter working memory, whether each unit uses text, image, or dual presentation, and how visual units are organized. All presentation decisions share one resource limit, $C_{\text{text}}+C_{\text{vis}}\leq B$.

\subsection{Query-Conditioned Evidence Extraction}

The existing evidence extraction methods based on memory passages or memory chunks are often too coarse-grained to directly support fine-grained memory management. For example, a document passage may contain only one sentence that is relevant to the query, while a long dialogue turn may include both key events and many irrelevant details. If the entire retrieved result is used as the memory context, such background content competes with truly useful facts for limited memory space. To address this issue, we use a query-conditioned evidence extractor that first locates useful content in the raw records, allowing subsequent steps to operate on smaller memory items with clearer semantics.

It outputs a set of source identifiers $\mathcal{S}$ for query-relevant memory chunks, together with character-level key spans $\mathcal{H}$ for each selected source. We denote this process as $(\mathcal{S},\mathcal{H})=E_{\phi}(q,\mathcal{R},B)$. Each key span contains its start and end positions within the source chunk. The source record and these spans jointly determine the content of each evidence unit. For each selected memory-chunk source, the system directly copies the corresponding content from the raw record and preserves the local context within a window of width $w$ around the key span. The resulting evidence unit is the smallest memory item that \memo can retain, reformat, or omit. It is defined as

\begin{equation}
\label{eq:unit_schema}
u_i=(\mathrm{id},x_i,h_i,s_i,c_i).
\end{equation}

The field $x_i$ stores the evidence content, $h_i$ stores the key span located by the extractor, and $s_i$ records the source. The field $c_i$ stores the priority rule for evidence retention. This rule depends on whether the content in $x_i$ belongs to the key span located by $h_i$ or to its associated context. It can require semantic information to be preserved in textual form, allow visual presentation to meet the budget constraint, or allow omission. As a result, subsequent modules can protect more critical semantic information when reducing memory.

\subsection{Evidence Extractor Training}

We construct extractor training examples from evidence annotations in the training sets of four benchmarks. We prioritize native evidence annotations from each benchmark and use deterministic weak labels when native span annotations are unavailable. In HotpotQA and 2WikiMultiHopQA, each context paragraph serves as a source chunk, and the official supporting facts $(\mathrm{title},\mathrm{sentence\ id})$ determine the selected paragraphs and supporting sentences. For LoCoMo, each dialogue turn with a speaker and a timestamp serves as a source chunk. The official evidence turn identifiers in each question-answer entry determine the selected sources, and the training, validation, and test sets are split at the dialogue level. For ALFWorld, we read tasks and high-level actions from the official expert trajectories, and deterministically replay the actions to construct adjacent pre-execution and post-execution states. Each entity in the pre-execution state serves as a source chunk. Entities that change in the next step are selected by state differencing, and the full serialized state block is labeled as state evidence.

Each training target contains selected source identifiers, annotated character spans, and mappings from sources to evidence units. Let $e^{\star}$ denote the serialized extraction target. We train the extractor through full-parameter supervised fine-tuning to generate this structured output from the query, candidate records, and budget using

\begin{equation}
\label{eq:extractor_sft}
\mathcal{L}_{\mathrm{ext}}(\phi)
=-\sum_t\log p_{\phi}(e_t^{\star}\mid e_{<t}^{\star},q,\mathcal{R},B).
\end{equation}

\subsection{Joint Evidence Selection and Presentation}

After evidence units are generated, the memory manager conditions on the current query, the evidence units, and the budget, and selects an action for each unit as its presentation form.
\begin{equation}
\label{eq:modes}
\mathcal{M}=\{\text{text},\text{image},\text{dual},\text{drop}\}.
\end{equation}

The action \texttt{drop} omits the unit from the current working memory. Omission saves resources, but it may also remove evidence needed for subsequent reasoning. For selected evidence units, the memory manager chooses different presentation forms according to their content. The action \texttt{text} preserves a compact textual anchor in the text package $T$, \texttt{image} places the unit into the visual page $V$, and \texttt{dual} both preserves the textual anchor and adds the unit to the visual structure. Text preserves precise semantic information, while images preserve temporal, tabular, or grouping relations. However, images consume the visual budget and may reduce the recoverability of strings. The dual form retains a precise textual anchor and adds structural information, with its cost counted toward both textual and visual resources. For visual structures, the memory manager further plans template preferences, which control whether visual units are rendered as cards, timelines, tables, checklists, or column layouts, so that different memory views can be constructed according to the meaning of the evidence. For example, in multi-hop questions, answer-related spans can be preserved as text, while relations from multiple documents are organized with cards. In long dialogues, key entities can be preserved as text, while event order is presented with a timeline. In embodied trajectories, goals and action parameters can be highlighted, while state changes are organized as a checklist.

Because content selection, presentation form, and visual organization compete for the same budget, the memory manager plans them jointly with a single evidence-unit-level plan. The plan contains the proposed action and confidence for each unit, the target memory usage, and the preferred page templates. After joint planning, the memory manager outputs the textual anchors and visual-structure information for the current query that are needed to construct the working memory.

\subsection{Memory Manager Training from Reader Outcomes}

We construct training examples from HotpotQA, 2WikiMultiHopQA, LoCoMo, and ALFWorld. For the same query and evidence, we try keeping one evidence unit as text, placing it in an image, keeping both forms, or leaving it out of the current memory. All other units remain fixed, which allows us to compare the effects of different counterfactual candidates. Each candidate memory is given to the same frozen reader to answer the question, and we record both the answer quality and the total memory usage. Candidates that violate the budget are removed. When multiple candidates yield similar answer quality, the one that uses fewer tokens is preferred. The final winning choice serves as the action label for that evidence unit.

Each action label is also associated with a confidence score. The score comes from the gap between the best action and the second-best action. Confidence is high when one presentation is clearly better and lower when two choices perform similarly. This value is scaled to a fixed range and only indicates how certain the action choice is for this training sample. Image and drop choices also pass a separate check based on actual reader results, preventing choices that may lose essential information from entering the training target.

The template label specifies how visual memory should be laid out. For each training sample, the template is determined by the answer quality of its counterfactual candidates. The training target stores an ordered list of candidate templates rather than a single template. During inference, the first template that satisfies the budget constraint is retained. The template preference is written into the same management plan together with the action and confidence.

The input of each training sample contains the query $q$, the evidence units $\mathcal{U}$, and the budget $B$, while the target $y^{\star}$ is the complete memory plan. The training objective of the memory manager is

\begin{equation}
\label{eq:sft}
\mathcal{L}_{\text{sft}}(\theta)
=-\sum_{t}\log p_{\theta}(y_t^{\star}\mid y_{<t}^{\star},q,\mathcal{U},B).
\end{equation}

\subsection{Working Memory Building}

\paragraph{Reading the memory plan.}

Working-memory construction takes the current query, the evidence units created by the extractor, the total budget, and the memory manager's plan as input. Each evidence unit contains content copied from the original memory together with its source and key passage. The plan gives a preferred presentation and confidence for every unit, as well as the target memory use and visual-template preferences. Under the shared budget, the builder decides which units enter the text packet, the visual page, both, or neither.

\paragraph{Preparing the text packet.}

Evidence units assigned text or dual become short text entries. The builder copies their key wording directly from the corresponding source records and keeps the unit and source identifiers with each entry, preserving the link to the original memory. The ordered entries form the text packet $T$. The unified text-token accounting rule gives its cost $C_{\text{text}}$, leaving the rest of the total budget for visual memory.

\paragraph{Preparing the visual page.}

Evidence units assigned image or dual become visual blocks containing a title, source, content, information type, and highlighted wording. The memory manager's template preferences arrange these blocks as a timeline, table, checklist, cards, or columns. Once the text cost is known, the remaining budget determines the available page size, and the same page size determines the visual cost $C_{\text{vis}}$. The renderer then lays out the blocks as the visual page $V$, with $C_{\text{text}}+C_{\text{vis}}\leq B$.

\paragraph{Assembling the reader input.}

The text packet and visual page together form the final working memory $a=(T,V)$. The system joins the text entries into readable text labeled by unit identifier and renders the visual page as an image. It then gives the current query, text, and image to the frozen reader, which produces an answer from this query-specific working memory. 

\begin{table*}[!t]
\centering
\small
\setlength{\tabcolsep}{5pt}
\begin{tabular}{@{}c|cc|cc|cc|cc|cc|cc@{}}
\toprule
\multirow{2}{*}{Reader} & \multirow{2}{*}{\begin{tabular}[c]{@{}c@{}}Memory\\ Type\end{tabular}} & \multirow{2}{*}{Method} & \multicolumn{2}{c|}{Overall} & \multicolumn{2}{c|}{2Wiki} & \multicolumn{2}{c|}{HotpotQA} & \multicolumn{2}{c|}{LoCoMo} & \multicolumn{2}{c}{ALFWorld} \\ \cmidrule(l){4-13} 
 &  &  & EM & F1 & EM & F1 & EM & F1 & EM & F1 & EM & F1 \\ \midrule
\multirow{9}{*}{\begin{tabular}[c]{@{}c@{}}InternVL3.5\\ 8B\end{tabular}} & None & No memory & 13.51 & 17.60 & 20.99 & 23.86 & 8.56 & 15.81 & 0.32 & 4.26 & 0.00 & 1.92 \\ \cmidrule(l){2-13} 
 & \multirow{4}{*}{Text} & Text-only & 32.93 & 45.36 & 36.66 & 47.84 & 43.53 & 58.59 & 16.56 & 43.37 & 1.37 & 10.34 \\
 &  & BM25 & 31.60 & 43.93 & 34.91 & 45.77 & 41.91 & 56.83 & 16.88 & 43.75 & \textbf{1.75} & \textbf{11.44} \\
 &  & Mem-$\alpha$ & 30.24 & 43.24 & 32.98 & 45.25 & 41.02 & 56.35 & \underline{18.40} & \underline{44.88} & 0.95 & 9.43 \\
 &  & MemAgent & 37.33 & 51.95 & 41.29 & \underline{55.98} & 50.97 & 68.35 & 13.53 & 39.31 & 0.00 & 6.87 \\ \cmidrule(l){2-13} 
 & \multirow{3}{*}{Visual} & Visual-only & 29.67 & 42.45 & 32.63 & 43.95 & 40.26 & 57.47 & 17.21 & 42.90 & 0.05 & 6.72 \\
 &  & AgentOCR & \underline{40.53} & \underline{52.60} & \underline{44.70} & 55.27 & \underline{54.58} & \textbf{71.02} & \textbf{22.40} & \textbf{49.12} & 0.71 & 6.63 \\
 &  & MemOCR & 16.08 & 22.08 & 24.07 & 28.24 & 11.45 & 20.88 & 1.95 & 10.51 & 0.33 & 5.24 \\ \cmidrule(l){2-13} 
 & Managed & MEMO (Ours) & \textbf{47.83} & \textbf{59.82} & \textbf{58.55} & \textbf{68.88} & \textbf{54.93} & \underline{70.09} & 17.21 & 43.05 & \underline{1.68} & \underline{10.62} \\ \midrule
\multirow{9}{*}{\begin{tabular}[c]{@{}c@{}}Qwen3-VL\\ 32B\end{tabular}} & None & No memory & 15.46 & 20.40 & 19.06 & 22.98 & 18.18 & 26.22 & 0.97 & 3.11 & 0.00 & 2.64 \\ \cmidrule(l){2-13} 
 & \multirow{4}{*}{Text} & Text-only & 40.23 & 52.99 & 45.59 & 56.26 & 51.16 & 66.90 & \textbf{26.41} & \textbf{51.20} & \textbf{1.90} & \textbf{13.82} \\
 &  & BM25 & 38.32 & 51.07 & 43.09 & 53.66 & 49.41 & 65.19 & 25.54 & \underline{50.50} & 1.53 & 13.59 \\
 &  & Mem-$\alpha$ & 41.35 & 52.31 & 47.09 & 56.05 & 53.16 & 67.96 & 24.46 & 48.99 & 0.46 & 8.23 \\
 &  & MemAgent & \underline{47.88} & \underline{62.00} & \underline{56.42} & \underline{69.80} & \textbf{58.16} & \textbf{75.38} & 23.92 & 48.71 & 1.61 & 10.20 \\ \cmidrule(l){2-13} 
 & \multirow{3}{*}{Visual} & Visual-only & 22.18 & 32.60 & 26.53 & 35.89 & 26.83 & 42.20 & 8.87 & 26.43 & 0.00 & 2.82 \\
 &  & AgentOCR & 32.88 & 45.54 & 38.57 & 48.87 & 41.02 & 60.39 & 13.64 & 37.93 & 0.00 & 5.21 \\
 &  & MemOCR & 20.61 & 28.06 & 27.04 & 32.08 & 21.37 & 33.49 & 2.27 & 12.14 & 0.05 & 5.90 \\ \cmidrule(l){2-13} 
 & Managed & MEMO (Ours) & \textbf{50.23} & \textbf{63.84} & \textbf{62.20} & \textbf{73.91} & \underline{55.83} & \underline{72.52} & \underline{25.76} & 50.22 & \underline{1.74} & \underline{13.81} \\ \midrule
\multirow{9}{*}{gpt-5.4-mini} & None & No memory & 20.40 & 22.98 & 27.17 & 28.51 & 20.63 & 26.67 & 0.32 & 0.45 & 0.00 & 0.27 \\ \cmidrule(l){2-13} 
 & \multirow{4}{*}{Text} & Text-only & 47.49 & 57.24 & 55.58 & 62.42 & 58.23 & 71.71 & 30.30 & \underline{54.99} & 0.71 & 10.45 \\
 &  & BM25 & 44.52 & 54.17 & 51.16 & 57.75 & 56.13 & 69.48 & 30.19 & 54.98 & 0.49 & \underline{10.67} \\
 &  & Mem-$\alpha$ & 46.34 & 55.65 & 54.12 & 60.95 & 57.29 & 70.58 & 29.98 & 53.41 & 0.07 & 7.45 \\
 &  & MemAgent & \underline{55.31} & \underline{65.81} & \underline{66.88} & \underline{75.28} & \textbf{65.20} & \textbf{80.16} & 26.52 & 51.40 & 0.20 & 6.49 \\ \cmidrule(l){2-13} 
 & \multirow{3}{*}{Visual} & Visual-only & 36.24 & 46.54 & 42.12 & 49.96 & 44.31 & 60.01 & 27.27 & 49.53 & 1.11 & 6.93 \\
 &  & AgentOCR & 41.74 & 52.75 & 47.38 & 55.75 & 53.28 & 68.80 & \textbf{30.95} & 54.31 & 0.73 & 9.63 \\
 &  & MemOCR & 28.96 & 37.53 & 35.11 & 41.42 & 33.69 & 46.20 & 10.50 & 26.67 & \underline{1.24} & 8.37 \\ \cmidrule(l){2-13} 
 & Managed & MEMO (Ours) & \textbf{57.78} & \textbf{68.62} & \textbf{71.68} & \textbf{80.06} & \underline{63.88} & \underline{78.14} & \underline{30.41} & \textbf{55.04} & \textbf{2.12} & \textbf{12.18} \\ \bottomrule
\end{tabular}
\caption{Results under 128-token budget limitation, reporting normalized exact match (EM) and token F1.}
\label{tab:fixed128-em-f1}
\end{table*}

\begin{table*}[!t]
\centering
\setlength{\tabcolsep}{5pt}
\small
\begin{tabular}{@{}c|cc|ccc|cc|cc|cc|cc@{}}
\toprule
\multirow{2}{*}{{Reader}} & \multirow{2}{*}{\begin{tabular}[c]{@{}c@{}}Memory\\ Type\end{tabular}} & \multirow{2}{*}{{Method}} & \multicolumn{3}{c|}{{Overall}} & \multicolumn{2}{c|}{{2Wiki}} & \multicolumn{2}{c|}{{HotpotQA}} & \multicolumn{2}{c|}{{LoCoMo}} & \multicolumn{2}{c}{{ALFWorld}} \\ \cmidrule(l){4-14} 
 &  &  & EM & F1 & Tokens $\downarrow$ & EM & F1 & EM & F1 & EM & F1 & EM & F1 \\ \midrule
\multirow{9}{*}{\begin{tabular}[c]{@{}c@{}}InternVL3.5\\ 8B\end{tabular}} & None & No   memory & 13.42 & 17.47 & 0.00 & 20.91 & 23.77 & 8.40 & 15.60 & 0.32 & 4.17 & 0.00 & 1.80 \\ \cmidrule(l){2-14} 
 & \multirow{4}{*}{Text} & Text-only & 48.99 & 60.82 & 253.89 & 57.52 & 67.49 & {\underline{60.60}} & {\underline{75.73}} & 18.18 & 44.90 & 1.21 & 10.29 \\
 &  & BM25 & \textbf{49.84} & {\underline{61.72}} & 253.89 & 58.57 & {\underline{68.58}} & \textbf{61.36} & \textbf{76.28} & 17.86 & 44.80 & \textbf{1.75} & \textbf{11.41} \\
 &  & Mem-$\alpha$ & 49.61 & 60.88 & 267.24 & {\underline{58.92}} & 68.07 & 60.33 & 75.42 & {\underline{20.45}} & {\underline{47.20}} & 0.71 & 8.94 \\
 &  & MemAgent & 45.17 & 58.88 & {\underline{170.23}} & 54.52 & 67.73 & 54.33 & 71.42 & 11.69 & 38.03 & 0.00 & 6.47 \\ \cmidrule(l){2-14} 
 & \multirow{3}{*}{Visual} & Visual-only & 36.65 & 50.75 & 4096.00 & 39.97 & 53.08 & 50.18 & 68.23 & 19.16 & 46.17 & 0.77 & 8.04 \\
 &  & AgentOCR & 42.29 & 53.94 & 177.02 & 47.06 & 56.88 & 56.33 & 72.87 & \textbf{22.40} & \textbf{47.36} & 0.71 & 6.52 \\
 &  & MemOCR & 45.86 & 57.59 & 705.66 & 52.58 & 62.57 & 58.98 & 74.67 & 19.48 & 46.00 & 0.55 & 7.72 \\ \cmidrule(l){2-14} 
 & Managed & MEMO   (Ours) & {\underline{49.78}} & \textbf{62.26} & \textbf{83.93} & \textbf{60.48} & \textbf{71.27} & 58.02 & 73.69 & 18.49 & 45.78 & {\underline{1.53}} & {\underline{10.75}} \\ \midrule
\multirow{9}{*}{\begin{tabular}[c]{@{}c@{}}Qwen3-VL\\ 32B\end{tabular}} & None & No memory & 16.15 & 20.98 & 0.00 & 20.17 & 23.80 & 18.55 & 26.71 & 0.97 & 3.27 & 0.00 & 2.64 \\ \cmidrule(l){2-14} 
 & \multirow{4}{*}{Text} & Text-only & 53.13 & 66.76 & 253.89 & 63.92 & 75.55 & 62.22 & {\underline{78.74}} & \textbf{26.95} & {\underline{53.16}} & {\underline{1.92}} & 14.45 \\
 &  & BM25 & \textbf{53.29} & \textbf{67.14} & 253.89 & 64.15 & 76.06 & \textbf{62.52} & \textbf{79.13} & \textbf{26.95} & 52.79 & 1.59 & {\underline{14.46}} \\
 &  & Mem-$\alpha$ & 52.33 & 64.96 & 267.35 & 62.90 & 74.26 & {\underline{62.25}} & 78.37 & 24.68 & 49.84 & 0.44 & 8.28 \\
 &  & MemAgent & 52.90 & 66.40 & {\underline{169.55}} & {\underline{65.12}} & \textbf{77.08} & 60.14 & 77.48 & 22.40 & 48.39 & 1.21 & 10.15 \\ \cmidrule(l){2-14} 
 & \multirow{3}{*}{Visual} & Visual-only & 41.86 & 59.12 & 4096.00 & 50.60 & 67.77 & 49.64 & 70.59 & 17.86 & 44.57 & 0.00 & 8.48 \\
 &  & AgentOCR & 43.01 & 56.02 & 177.02 & 51.43 & 62.00 & 51.55 & 70.57 & 22.40 & 48.14 & 0.16 & 7.20 \\
 &  & MemOCR & 51.30 & 64.20 & 705.66 & 62.61 & 72.83 & 59.84 & 78.25 & 20.13 & 46.05 & 0.27 & 9.07 \\ \cmidrule(l){2-14} 
 & Managed & MEMO   (Ours) & {\underline{53.27}} & {\underline{66.77}} & \textbf{83.93} & \textbf{65.92} & {\underline{77.06}} & 59.30 & 76.11 & {\underline{26.62}} & \textbf{53.27} & \textbf{1.97} & \textbf{14.61} \\ \midrule
\multirow{9}{*}{gpt-5.4-mini} & None & No memory & 20.45 & 23.05 & 0.00 & 27.18 & 28.70 & 20.77 & 26.57 & 0.32 & 0.44 & 0.00 & 0.28 \\ \cmidrule(l){2-14} 
 & \multirow{4}{*}{Text} & Text-only & 58.43 & 69.29 & 253.89 & 71.34 & 79.41 & 67.16 & 82.05 & 31.49 & {\underline{57.05}} & 0.77 & 10.57 \\
 &  & BM25 & \textbf{58.94} & 69.74 & 253.89 & {\underline{72.28}} & 80.17 & 67.43 & 82.39 & 30.52 & 55.67 & 0.55 & 10.50 \\
 &  & Mem-$\alpha$ & 57.56 & 67.98 & 267.97 & 70.07 & 78.19 & 66.86 & 81.59 & 30.52 & 54.86 & 0.16 & 7.43 \\
 &  & MemAgent & 57.34 & 67.91 & {\underline{169.65}} & 70.12 & 78.52 & 66.30 & 81.67 & 26.62 & 51.99 & 0.33 & 6.15 \\ \cmidrule(l){2-14} 
 & \multirow{3}{*}{Visual} & Visual-only & 58.62 & \textbf{70.27} & 4096.00 & 71.72 & {\underline{80.88}} & {\underline{67.51}} & \textbf{83.04} & 29.87 & 56.06 & 0.27 & 10.17 \\
 &  & AgentOCR & 48.48 & 60.02 & 177.02 & 55.69 & 64.22 & 60.79 & 76.51 & \textbf{32.79} & 55.32 & 1.32 & {\textbf{12.90}} \\
 &  & MemOCR & 57.76 & 68.71 & 705.66 & 69.58 & 78.48 & \textbf{67.65} & {\underline{82.52}} & 29.55 & 53.63 & {\underline{1.75}} & 9.61 \\ \cmidrule(l){2-14} 
 & Managed & MEMO   (Ours) & {\underline{58.86}} & {\underline{69.84}} & \textbf{83.93} & \textbf{73.06} & \textbf{81.48} & 65.02 & 79.52 & {\underline{31.80}} & \textbf{57.58} & \textbf{1.99} & \underline{12.20} \\ \bottomrule
\end{tabular}
\caption{Results without source-memory truncation under the 4,096-token execution window, reporting normalized exact match (EM) and token F1.}
\label{tab:untruncated4096-em-f1}
\end{table*}

\section{Experiments}

\subsection{Implementation Details}

Both trained components, the evidence extractor and the memory manager, use Qwen2.5-1.5B-Instruct as their backbone. We conducted supervised fine-tuning with \texttt{verl} \cite{sheng2024hybridflow}. The reader used to train memory manager is InternVL3.5-4B-Instruct. All experiments were conducted on NVIDIA A100 GPUs.

\subsection{Evaluation Settings}

The HotpotQA and 2WikiMultiHopQA test sets come from official development sets with reference answers, each pre-split into non-overlapping validation and test sets. The LoCoMo test set contains two fully held-out conversations. ALFWorld test samples are next-action prediction instances derived from official valid\_unseen expert trajectories. HotpotQA and 2WikiMultiHopQA evaluate multi-hop document reading, LoCoMo evaluates memory over long conversations, and ALFWorld evaluates whether task and state information retained in memory supports recovering the expert's next action. This design gives all four benchmarks the same Reader output format and metrics, enabling a clear, consistent Overall score.

\paragraph{Baselines.}

We compare MEMO with three direct controls and five published baselines. No memory gives the Reader only the query. Text-only and Visual-only present the available memory directly as text or a rendered page, without a learned memory constructor. BM25 ranks textual evidence by lexical relevance to the query \cite{robertsonProbabilisticRelevanceFramework2009}. The learned text-memory baselines are Mem-$\alpha$, which learns memory construction through reinforcement learning \cite{wangMemaLearningMemory2025}, and MemAgent, which reads the source in segments and repeatedly updates a compact text memory \cite{yuMemAgentReshapingLongContext2025}. The visual-memory baselines are AgentOCR, which compresses history into OCR-style images \cite{fengAgentOCRReimaginingAgent2026}, and MemOCR, which constructs layout-aware visual memory \cite{shiMemOCRLayoutAwareVisual2026}. Each resulting memory is passed to the same frozen Reader so that the comparison reflects how the memory is constructed.

\paragraph{Readers.}

We evaluate every method with three frozen multimodal Readers: InternVL3.5-8B, Qwen3-VL-32B, and gpt-5.4-mini. They cover different model scales and families, allowing us to test whether a memory method depends on one particular backend. For a given Reader, all methods receive the same query and source memory, and the Reader remains unchanged throughout evaluation.

\paragraph{Memory budgets.}

We compare a strict 128-token budget with a setting that imposes no token budget. Under 128-token budget, every method must satisfy $\text{text tokens}+\text{visual tokens}\leq128$, enabling a direct quality comparison under the same tight limit. In the no-budget setting, we impose no method-level memory limit and do not truncate the source memory. Execution still uses a 4,096-token LLM context window, but the complete source memory of every test example fits within it. No method is therefore forced to discard or compress content to meet an experimental cap, so this setting is effectively unrestricted.

\paragraph{Metrics.}

The result tables report normalized exact match (EM) and token-level F1 as percentages. After converting all answers to lowercase and removing punctuation and English articles, EM assigns a score only when the prediction exactly matches the reference answer, whereas F1 gives partial credit based on token overlap between the two answers. EM measures strict correctness, while F1 is more sensitive to partially correct content in multi-token answers. Using the same two metrics also allows a unified Overall score to be computed across the four tasks. For ALFWorld, we therefore evaluate whether the Reader can predict the expert's next action from the task and the saved trajectory state, and compare the prediction with the reference action.

\begin{table}[!t]
\centering
\small
\begin{tabular}{@{}cc|ccc@{}}
\toprule
\multirow{2}{*}{\begin{tabular}[c]{@{}c@{}}Evidence\\ Extractor\end{tabular}} & \multirow{2}{*}{\begin{tabular}[c]{@{}c@{}}Memory\\ Manager\end{tabular}} & \multicolumn{3}{c}{Overall} \\ \cmidrule(l){3-5} 
 &  & EM & F1 & Tokens \\ \midrule
\ding{55} & \ding{55} & 47.49 & 57.24 & 126.22 \\
\ding{51} & \ding{55} & 53.56 & 66.09 & 70.12 \\
\ding{51} & \ding{51} & 57.78 & 68.62 & 81.07 \\ \bottomrule
\end{tabular}
\caption{Component ablation under the 128-token budget with gpt-5.4-mini. Tokens denotes average memory use.}
\label{tab:component-ablation-fixed128-gpt54mini}
\end{table}

\subsection{Main Results}

Tables~\ref{tab:fixed128-em-f1} and~\ref{tab:untruncated4096-em-f1} compare MEMO with text-only, visual-only, and other memory methods under a tight 128-token budget and with no token budget.

\paragraph{128-token limit.}
Under the strict 128-token limit, MEMO achieves the best Overall EM and F1 with all three readers. With InternVL3.5-8B, it improves over the strongest baseline by 7.30 EM points and 7.22 F1 points. The gains remain clear with the larger readers: MEMO leads the strongest baseline by 2.35 EM and 1.84 F1 points with Qwen3-VL-32B, and by 2.47 EM and 2.81 F1 points with gpt-5.4-mini. The improvement is therefore not tied to one particular reader. MEMO is especially strong on 2Wiki, where it ranks first in both metrics for every reader, and it remains first or second on HotpotQA. Although another baseline is better on some individual LoCoMo and ALFWorld scores, MEMO keeps the clearest advantage when all test examples are considered together.

\paragraph{No token budget.}
When no token budget is imposed, MEMO uses only 83.93 tokens on average for every reader. This is 50.5--50.7\% fewer tokens than MemAgent, the next smallest nonzero-memory baseline, and 66.9\% fewer than Text-only. Even with this lower memory use, MEMO remains near the top in overall answer quality. It obtains the best Overall F1 with InternVL3.5-8B and ranks second on its Overall EM by only 0.06 points. With Qwen3-VL-32B and gpt-5.4-mini, it is second on both Overall metrics, and its largest gap from the best result is only 0.43 points. Thus, MEMO is first or second on every Overall EM and F1 result while using the least memory among all methods that provide memory.

\paragraph{Overall takeaway.}
The two settings show complementary advantages. Under a hard limit, MEMO gets more useful information into the same small budget and produces better answers. When much more context is available, it does not spend tokens simply because it can. It preserves near-best answer quality with substantially less memory. The same pattern across three different readers suggests that the advantage comes from MEMO's memory management rather than from a favorable match to one reader.

\subsection{Component Ablation}

Table~\ref{tab:component-ablation-fixed128-gpt54mini} studies the two trainable components under the same 128-token budget with gpt-5.4-mini.

With both components disabled, the system reaches 47.49 EM and 57.24 F1 using 126.22 memory tokens. Enabling only the evidence extractor raises EM by 6.07 points and F1 by 8.85 points. Meanwhile, memory use falls to 70.12 tokens, a 44.4\% reduction. This shows that selecting query-relevant evidence can improve the reader's answer while removing substantial unnecessary memory.

Adding the memory manager on top of the evidence extractor further raises EM from 53.56 to 57.78 and F1 from 66.09 to 68.62, gains of 4.22 and 2.53 points. Memory use rises moderately from 70.12 to 81.07 tokens because the memory manager keeps a slightly richer working memory, but remains well below the 126.22 tokens used when both components are disabled. This shows that the memory manager does not simply minimize token count: it spends a few additional tokens when they improve answers.

Compared with disabling both components, the complete system improves EM by 10.29 points and F1 by 11.38 points while using 35.8\% fewer memory tokens. The extractor provides the first large improvement, and the memory manager adds a further gain, so the two modules make distinct, complementary contributions to the final result.

\section{Conclusion}

We propose MEMO, a query-conditioned memory readout method that converts memory into evidence units and jointly selects content, textual or visual presentation, and layout under a token budget. Across four benchmarks, MEMO achieves the best overall EM and F1 under the 128-token constraint when using all three readers. Under the 4,096-token window, it uses 83.93 tokens on average while ranking first or second on every overall metric. Ablation studies show that evidence extraction provides the main gains in compression and accuracy by preserving query-relevant information before presentation choices, while the memory manager further improves quality. These results indicate that the proposed evidence-level multimodal organization serves as an effective and efficient memory interface.

\bibliography{aaai2027}

\end{document}